%% file: main.tex
\documentclass[twoside,11pt]{article}

\usepackage{jmlr2e}

\jmlrheading{\smiley}{2022}{\smiley}{01/22}{\smiley/\smiley}{\smiley}{Buyun Liang and Tim Mitchell and Ju Sun}

\ShortHeadings{NCVX}{Liang and Mitchell and Sun}
\firstpageno{1}

\usepackage{graphicx}

\newcommand{ \paren }[1]{ \ensuremath {\left(  #1 \right)} }
\newcommand{\norm}[2]{\left\| #1 \right\|_{#2}}

\usepackage{algorithm}
\usepackage{algpseudocode}
\usepackage{xcolor}
\usepackage{amsmath,amsfonts,amssymb,amsthm, bm}
\usepackage{multicol}
\usepackage{wasysym}

\let\oldhref\href
\renewcommand{\href}[2]{\oldhref{#1}{\bfseries#2}}
\newcommand{\ncvx}{\texttt{NCVX}\ }
\newcommand{\granso}{\texttt{GRANSO}\ }
\newcommand{\pygranso}{\texttt{PyGRANSO}\ }

\input{math_commands.tex}

\newcommand{\T}{\intercal}

\usepackage{listings}
\usepackage{color}

\definecolor{dkgreen}{rgb}{0,0.6,0}
\definecolor{gray}{rgb}{0.5,0.5,0.5}
\definecolor{mauve}{rgb}{0.58,0,0.82}

\begin{document}

\title{\texttt{NCVX}: A User-Friendly and Scalable Package for Nonconvex Optimization in Machine Learning}

\author{\name Buyun Liang$^1$ \email liang664@umn.edu \\
        \name Tim Mitchell$^2$ \email mitchell@mpi-magdeburg.mpg.de\\
       \name Ju Sun$^1$ \email jusun@umn.edu \\
       \\
       \addr $^1$ Department of Computer Science \& Engineering, 
       University of Minnesota, Twin Cities, USA\\
       \addr $^2$ Max Planck Institute for Dynamics of Complex Technical Systems, Germany}

\editor{\smiley\ and \smiley}

\maketitle

\begin{abstract}
Optimizing nonconvex (NCVX) problems, especially nonsmooth and constrained ones, is an essential part of machine learning. However, it can be hard to reliably solve such problems without optimization expertise. Existing general-purpose NCVX optimization packages are powerful but typically cannot handle nonsmoothness. \granso is among the first optimization solvers targeting general nonsmooth NCVX problems with nonsmooth constraints, but, as it is 
implemented in MATLAB and requires the user to provide analytical gradients, 
\granso is often not a convenient choice in machine learning (especially deep learning) applications.
To greatly lower the technical barrier, 
we introduce a new software package called \texttt{NCVX}, whose initial release
contains the solver \texttt{PyGRANSO}, a PyTorch-enabled port of \granso 
incorporating  auto-differentiation, GPU acceleration, tensor input, and support for new QP solvers.
\ncvx is built on freely available and widely used open-source frameworks, and as a highlight, can solve general constrained deep learning problems, the first of its kind.
\texttt{NCVX} is available at \url{https://ncvx.org}, with detailed documentation and numerous examples from machine learning and other fields. 
\end{abstract}


\section{Introduction}

Mathematical optimization is an indispensable modeling and computational tool for all science and engineering fields, especially for machine learning. To date, researchers have developed numerous foolproof techniques and user-friendly solvers and modeling languages for convex (CVX) problems, such as SDPT3~\citep{toh1999sdpt3}, Gurobi~\citep{gurobi}, Cplex~\citep{cplex2009v12}, TFOCS~\citep{becker2011templates}, 
CVX(PY)~\citep{grant2008cvx,diamond2016cvxpy}, AMPL~\citep{gay2015ampl}, YALMIP~\citep{LofbergYALMIP}. These developments have substantially lowered the barrier of CVX optimization for non-experts. However, practical problems, especially from machine and deep learning, are often nonconvex (NCVX), and possibly also nonsmooth (NSMT) and constrained (CSTR). 

There are methods and packages handling NCVX problems in restricted settings: 
PyTorch~\citep{paszke2019pytorch} and TensorFlow~\citep{tensorflow2015-whitepaper} can solve large-scale NCVX, NSMT problems without constraints. CSTR problems can be heuristically turned into penalty forms, thus making the constraints implicit,
but this may not produce feasible solutions for the original problems. When the constraints are simple, structured methods such as projected (sub)gradient and Frank-Wolfe~\citep{sra2012optimization} can be used. When the constraints are differentiable manifolds, one can consider manifold optimization methods and packages, e.g., (Py)manopt~\citep{boumal2014manopt,townsend2016pymanopt}, Geomstats~\citep{miolane2020geomstats}, McTorch~\citep{MeghwanshiEtAl2018McTorch}, and Geoopt~\citep{KochurovEtAl2020Geoopt}. For general CSTR problems, KNITRO~\citep{pillo2006large} and IPOPT~\citep{WaechterBiegler2005implementation} implement interior-point methods, while ensmallen~\citep{curtin2021ensmallen} and GENO~\citep{laue2019geno} rely on augmented Lagrangian methods. However, 
moving beyond smooth (SMT) constraints, both of these families of methods, at best, 
handle only special types of NSMT constraints. Finally, packages specialized for machine learning, such as scikit-learn~\citep{pedregosa2011scikit}, MLib~\citep{meng2016mllib} and Weka~\citep{witten2005practical}, often use problem-specific solvers that cannot be easily extended to new formulations. 

\section{The \granso and \ncvx packages}

\texttt{GRANSO}\footnote{\url{http://www.timmitchell.com/software/GRANSO/}} is among the first optimization packages that can handle general NCVX, NSMT, CSTR problems~\citep{curtis2017bfgs}:
\begin{align}
\begin{split}
     & \min_{\vx \in \R^n} f(\vx),\;\;\text{s.t. } c_i(\vx) \leq 0,\; \forall\; i \in \mathcal I;\;\; c_i(\vx)=0,\; \forall\; i \in \mathcal E. 
\end{split}
\end{align}
Here, the objective $f$ and constraint functions $c_i$'s are only required to be almost everywhere continuously differentiable. \granso is based on quasi-Newton updating 
with sequential quadratic programming (BFGS-SQP) and has the following advantages: (1) \textbf{unified treatment of NCVX problems}: no need to distinguish CVX vs NCVX and SMT vs NSMT problems, similar to typical nonlinear programming packages;  (2) \textbf{reliable step-size rule}: specialized methods for NSMT problems, such as subgradient and proximal methods, often entail tricky step-size tuning and require the expertise to recognize the structures~\citep{sra2012optimization}, while
\granso chooses step sizes adaptively via a gold-standard line search; (3) \textbf{principled stopping criterion}: \granso stops its iteration by checking a theory-grounded stationarity condition for NMST problems, whereas specialized methods are usually stopped when reaching ad-hoc iteration caps. 

However, \granso users must derive gradients analytically\footnote{\granso is implemented in MATLAB and does not support auto-differentiation, although recent versions of MATLAB have included primitive auto-differentiation functionalities. } and then provide code for these computations,
a process which may require some expert knowledge, is often error-prone, and 
in machine learning, is generally impractical, e.g., for the training of large neural networks. Furthermore, as part of the MATLAB\ software ecosystem, 
\granso is generally hard for practitioners to integrate it with existing popular machine learning frameworks---mostly in Python and R---and users' own existing toolchains.
To overcome these issues and facilitate both high performance and ease of use in machine and deep learning, 
we introduce a new software package called \texttt{NCVX}, whose initial release contains the solver
\texttt{PyGRANSO}, a PyTorch-enabled port of \granso with several new key features:
(1) auto-differentiation of all gradients, which is a main feature to make \pygranso user-friendly; 
(2) support for both CPU and GPU computations for improved hardware acceleration and massive parallelism; 
(3) support for general tensor variables including vectors and matrices, 
   as opposed to the single vector of concatenated optimization variables that \granso uses;  
(4) integrated support for OSQP~\citep{osqp} and other QP solvers 
for respectively computing search directions and the stationarity measure on each iteration. 
OSQP generally outperforms commercial QP solvers in terms of scalability and speed. 
All of these enhancements are crucial for solving large-scale machine learning problems. 
\texttt{NCVX}, licensed under the AGPL version 3,  is built entirely on freely available and widely used open-source frameworks; see \url{https://ncvx.org} for documentation and examples.

\section{Usage Examples: Dictionary Learning and Neural Perceptual Attack}

In order to make \ncvx friendly to non-experts, we strive to keep the user input minimal. The user is only required to specify the optimization variables (names and dimensions of variables) and define the objective and constraint functions. Here, we briefly demonstrate the usage of \pygranso solver on a couple of machine learning problems. 


\paragraph{Orthogonal Dictionary Learning (ODL, \cite{bai2018subgradient})} 

\begin{figure}[!htbp]
    \centering 
    \includegraphics[width=0.96\textwidth]{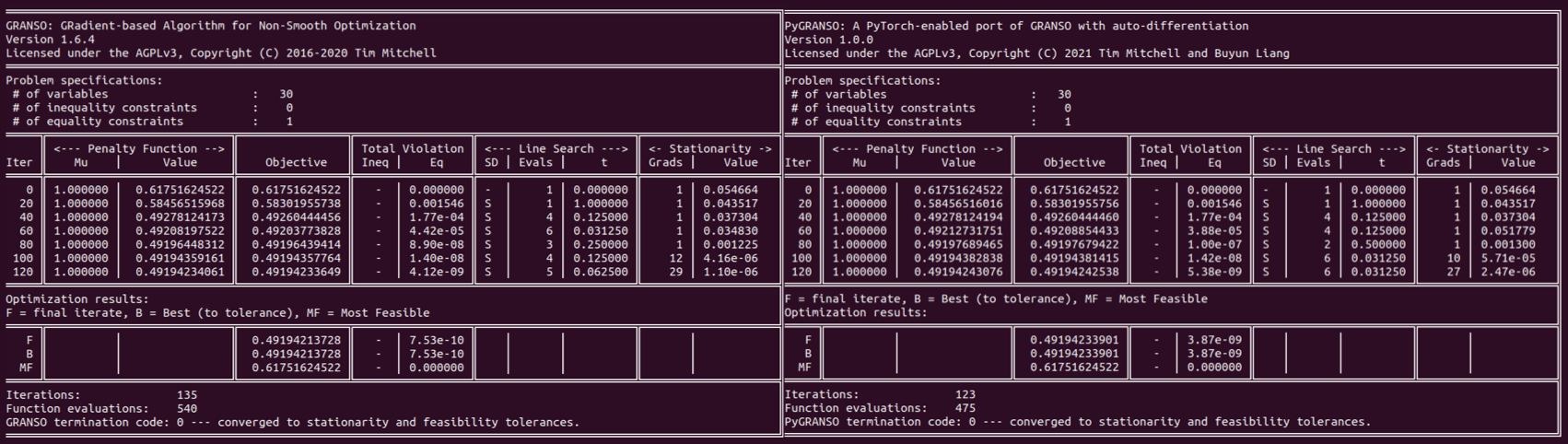}
    \caption{Consistency of \granso (left) and \pygranso (right) on ODL}
    \label{fig:consistency}
\end{figure}

One hopes to find a ``transformation'' $\vq \in \R^n$ to sparsify a data matrix $\mY \in \R^{n \times m}$: 
\begin{align} \label{eq:odl}
        \min_{\vq \in \R^n}\; f(\vq) \doteq \frac{1}{m} \norm{\vq^\T \mY}{1}, \quad \text{s.t.} \; \norm{\vq}{2} = 1,
\end{align}
where the sphere constraint $\norm{\vq}{2} = 1$ is to avoid the trivial solution $\vq = \bm 0$. Problem (\ref{eq:odl}) is NCVX, NSMT, and CSTR: nonsmoothness comes from the objective, and nonconvexity comes from the constraint. Demo \ref{demo1} and Demo \ref{demo2} show the implementations of ODL in \granso and \texttt{PyGRANSO}, respectively. Note that the analytical gradients of the objective and constraint functions are not required in \texttt{PyGRANSO}. Figure \ref{fig:consistency} shows that \pygranso produces results that are faithful to that of \granso on ODL. 
\begin{multicols}{2}
\begin{lstlisting}[language=matlab, caption={\granso for ODL},captionpos=b,label={demo1}]
function[f,fg,ci,cig,ce,ceg]=comb_fn(q)
    f = 1/m*norm(q'*Y, 1); % obj
    fg = 1/m*Y*sign(Y'*q); % obj grad
    ci = []; cig = []; % no ineq constr
    ce = q'*q - 1; % eq constr
    ceg = 2*q; % eq constr grad
end
soln = granso(n,comb_fn);
\end{lstlisting}
\columnbreak

\begin{lstlisting}[caption={\pygranso for ODL},captionpos=b,label={demo2}]
def comb_fn(X_struct):
    q = X_struct.q
    f = 1/m*norm(q.T@Y, p=1) # obj
    ce = pygransoStruct()
    ce.c1 = q.T@q - 1 # eq constr
    return [f,None,ce]
var_in = {"q": [n,1]} # define variable
soln = pygranso(var_in,comb_fn)
\end{lstlisting}
\end{multicols}
\paragraph{Neural Perceptual Attack (NPA, \cite{laidlaw2020perceptual})} The CSTR deep learning problem, NPA, is shown below:
\begin{align}\label{f3}
    \max_{\widetilde{\vx}}\;  \gL \paren{f\paren{\widetilde{\vx}},y},\;\;\text{s.t.}\;\; d\paren{\vx,\widetilde{\vx}} = \norm{\phi\paren{\vx} - \phi \paren{\widetilde{\vx}} }{2} \leq \epsilon. 
\end{align}
Here, $\vx$ is an input image, and the goal is to find its perturbed version $\widetilde{\vx}$ that is perceptually similar to $\vx$ (encoded by the constraint) but can fool the classifier $f$ (encoded by the objective). The loss $\gL\paren{\cdot,\cdot}$ is the margin loss used in~\citet{laidlaw2020perceptual}. Both $f$ in the objective and $\phi$ in the constraint are deep neural networks with ReLU activations, making both the objective and constraint functions NSMT and NCVX. The $d\paren{\vx,\widetilde{\vx}}$ distance is called the Learned Perceptual Image Patch Similarity (LPIPS)~\citep{laidlaw2020perceptual,zhang2018unreasonable}. Demo \ref{demo3} is the \pygranso example for solving problem (\ref{f3}). Note that the codes for data loading, model specification, loss function, and LPIPS distance are not included here. It is almost impossible to derive analytical subgradients for problem (\ref{f3}), and thus the auto-differentiation feature in \pygranso is necessary for solving it.

\begin{lstlisting}[caption={\pygranso for NPA},captionpos=b,label={demo3}]
def comb_fn(X_struct): 
    adv_inputs = X_struct.x_tilde 
    f = MarginLoss(model(adv_inputs),labels) # obj
    ci = pygransoStruct()
    ci.c1 = lpips_dists(adv_inputs) - 0.5  # ineq constr. percep bound epsilon=0.5
    return [f,ci,None] # No eq constr
var_in = {"x_tilde": list(inputs.shape)} # define variable
soln = pygranso(var_in,comb_fn)
\end{lstlisting}

\section{Roadmap}
Although \texttt{NCVX}, with the \pygranso solver, already has many powerful features, we plan to further improve it by adding several major components: (1) \textbf{symmetric rank one (SR1)}: SR1, another major type of quasi-Newton methods, allows less stringent step-size search and tends to help escape from saddle points faster by taking advantage of negative curvature directions~\citep{dauphin2014identifying};  (2) \textbf{stochastic algorithms}: in machine learning, computing with large-scale datasets often involves finite sums with huge number of terms, calling for stochastic algorithms for reduced per-iteration cost and better scalability~\citep{sun2019optimization}; (3) \textbf{conic programming (CP)}:  semidefinite programming and second-order cone programming, special cases of CP, are abundant in machine learning, e.g., kernel machines~\citep{zhang2019conic}; (4) \textbf{minimax optimization (MMO)}:  MMO is an emerging modeling technique in machine learning, e.g., generative adversarial networks (GANs) \citep{goodfellow2020generative} and multi-agent reinforcement learning \citep{jin2020local}. 






\acks{We would like to thank Frank E. Curtis and Michael L. Overton for their involvement in creating the BFGS-SQP algorithm that is implemented in the software package \texttt{GRANSO}. This work was supported by UMII Seed Grant Program and NSF CMMI 2038403.} 

\bibliography{jmlr}

\end{document}

%% file: math_commands.tex
\usepackage{amsmath,amsfonts,bm}

\def\eqref#1{equation~\ref{#1}}

\def\1{\bm{1}}

\def\vq{{\bm{q}}}

\def\vx{{\bm{x}}}

\def\mY{{\bm{Y}}}

\DeclareMathAlphabet{\mathsfit}{\encodingdefault}{\sfdefault}{m}{sl}
\SetMathAlphabet{\mathsfit}{bold}{\encodingdefault}{\sfdefault}{bx}{n}

\def\gL{{\mathcal{L}}}

\newcommand{\R}{\mathbb{R}}

